%% file: root.tex
\documentclass[letterpaper, 10 pt, conference]{ieeeconf}  
\usepackage{cite}
\usepackage{amsmath,amssymb,amsfonts}
\usepackage{graphicx}
\usepackage{textcomp}
\usepackage{multirow}
\usepackage{multicol}
\usepackage[dvipsnames]{xcolor}
\usepackage{hyperref}
\usepackage{placeins}
\usepackage{soul}
\usepackage{algorithm}
\usepackage{algpseudocode}
\usepackage{amssymb}
\algnewcommand\algorithmicforeach{\textbf{for each}}
\algdef{S}[FOR]{ForEach}[1]{\algorithmicforeach\ #1\ \algorithmicdo}

\title{\LARGE \bf
Expanding Versatility of Agile Locomotion through Policy Transitions Using Latent State Representation
}

\begin{document}

\author{Guilherme Christmann*, Ying-Sheng Luo*, Jonathan Hans Soeseno*, Wei-Chao Chen \\
        Inventec Corporation, Taipei, Taiwan \\ 
        \textit{\{guilherme.christmann, luo.ying-sheng, soeseno.jonathan, chen.wei-chao\}@inventec.com}}

\newcommand{\ct}[1]{\textcolor{Blue}{#1}}
\newcommand{\jt}[1]{\textcolor{magenta}{#1}}

\maketitle

\begingroup\renewcommand\thefootnote{*}
\footnotetext{These authors contributed equally, listed alphabetically by last name.}
\endgroup

\thispagestyle{empty}
\pagestyle{empty}

\begin{abstract}
This paper proposes the \textit{transition-net}, a robust transition strategy that expands the versatility of robot locomotion in the real-world setting. To this end, we start by distributing the complexity of different gaits into dedicated locomotion policies applicable to real-world robots. Next, we expand the versatility of the robot by unifying the policies with robust transitions into a single coherent meta-controller by examining the latent state representations. Our approach enables the robot to iteratively expand its skill repertoire and robustly transition between any policy pair in a library. In our framework, adding new skills does not introduce any process that alters the previously learned skills. Moreover, training of a locomotion policy takes less than an hour with a single consumer GPU. Our approach is effective in the real-world and achieves a 19\% higher average success rate for the most challenging transition pairs in our experiments compared to existing approaches.


\end{abstract}

\input{Sections/01_introduction.tex}

\input{Sections/02_related_works.tex} 
\input{Sections/03_method.tex}

\input{Sections/04_implementation_details.tex}
\input{Sections/05_experiments.tex} 
\input{Sections/06_conclusion.tex}



\section*{Acknowledgement}
We wish to thank Trista Chen for the meaningful discussions that lead to the design choices of our proposed work and our colleagues from the Inventec AI Center for their help in producing the supplementary video. 

\bibliographystyle{IEEEtran}
\bibliography{bibliography}

\end{document}

%% file: Sections/01_introduction.tex
\section{Introduction}
Robotics is impactful in many applications, from friendly human companions to industrial robots for automation and specialized robots in exploration industries. All these applications require robust and versatile movements to complete the assigned tasks. For instance, when exploring challenging terrains, the robot needs to handle unexpected disturbances and be versatile to maneuver through obstacles. These properties are enabled by locomotion policies that coherently choreograph every joint of the robot. However, as the skillset of the robot expands, it becomes exponentially intricate to maintain movement robustness.

To tackle this problem, it is common to develop an initial locomotion policy in simulation using deep reinforcement learning. This initial policy can be optimized for various objectives such as energy consumption \cite{fu2021minimizing}, task-based objectives \cite{iscen2021learning}, and imitating motions from a reference animation clip \cite{peng2018deepmimic}. Next, these policies are transferred to the real-world setting through domain randomization \cite{tobin2017domain}, and domain adaptation \cite{peng2020learning} methods. Although this setup can produce agile locomotion policies in the real-world, they struggle with expanding the skill repertoire of the robot. It is because the setup requires the re-training or fine-tuning of the initial policy in simulation and repeating the same transfer process. This modification further causes intricacies as the added skills impose varying complexity and require the delicate design of reward functions.

We adopt an alternative setup, where the complexity of each skill is contained into independent policies \cite{soeseno2021transition}, with flexibility of the architecture and objective function. The skill repertoire is represented by a library of policies which can include high-level directives \cite{luo2020carl, peng2021amp}. We start by training robust policies that can be transferred to a real quadruped. Next, to expand the versatility of the robot, we introduce a transition mechanism between the policies. As pointed out by \cite{starke2020localphase, soeseno2021transition, holden2017phase, zhang2018mann, starke2019nsm, peng2018deepmimic}, it is possible to use the phase variable as a low-dimensional proxy for the state of the agent. While this approximation can be successful in simulation, it is less accurate in a real-world setting since the same phase value can represent two significantly different states of the agent. 

\begin{figure}[t!]
    \centering
    \includegraphics[width=0.48\textwidth]{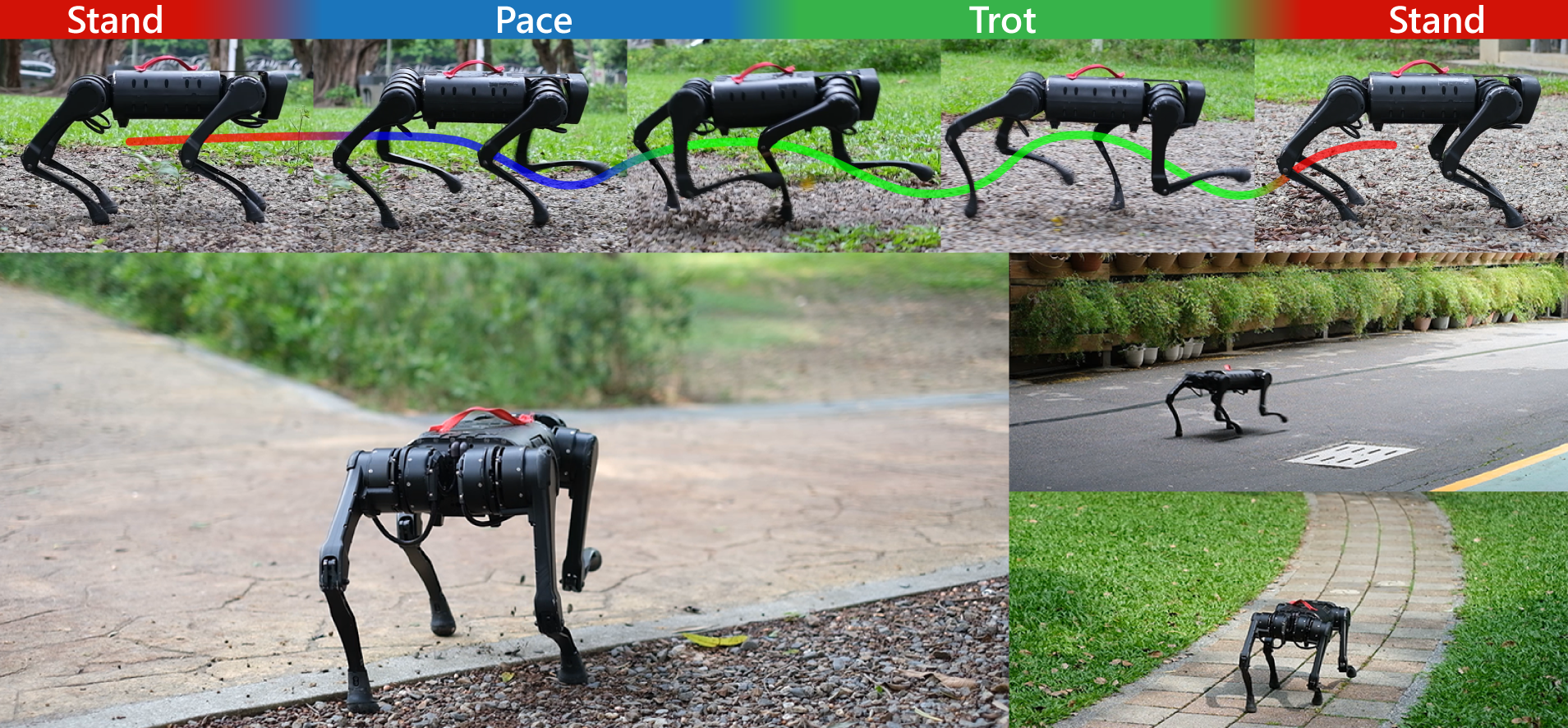}
    \caption{Our meta-controller enables the transition between locomotion policies in various real-world environments. We adopt three representative gaits within the robot's capabilities to perform static-to-dynamic, dynamic-to-dynamic, and dynamic-to-static transitions.}
    \label{fig:teaser}
\end{figure}

In our work, we use the latent state representations of the policies to encode the state of the agent and its environment conditions. We design a robust learning-based transition strategy that identifies successful transitions across all policy pairs. We then compose a meta-controller that regulates transitions between policies acting on the robot (see Figure~\ref{fig:teaser}). We summarize our contributions as follows,
\begin{itemize}
    \item A robust transition strategy, \textit{transition-net} that can accurately identify viable transitions between policies in the real-world using latent state representations;
    \item A scheme to iteratively expand the policy library of the robot without altering existing policies; and
    \item A demonstration and evaluation of the proposed method with a real-world quadruped robot.
\end{itemize}

%% file: Sections/02_related_works.tex
\section{Related Works}
\par \emph{\textbf{Non-Learning Controllers}}. Legged locomotion for robotics is a field with a large body of work that aims to emulate the efficiency and agility of the locomotion gaits observed in nature \cite{alexander1984gaits, hoyt1981gait}. A legged robot is able to move by producing a periodic sequence of foot contacts with the ground, called a gait \cite{haynes2006gaits}. For a long time, gaits have been produced via analytical models of the dynamics of a legged system \cite{wieber2016modeling}. Traditional approaches include using central pattern generators (CPG) to produce an oscillatory motion pattern that is further optimized into a desired trajectory \cite{fukuoka2013analysis, ijspeert2008central, barasuol2013reactive}. Model predictive control (MPC) is another approach used to generate optimized trajectories resulting in various gaits \cite{farshidian2017real, di2018dynamic, ding2019real, neunert2018whole}. While many works focus on optimizing a single gait, others also made efforts toward optimizing multiple gaits and enabling transitions \cite{haynes2011gait, nansai2015novel, boussema2019online}. 

\par \emph{\textbf{Learning-Based Controllers}}. Recently, deep reinforcement learning has become a popular approach to tackle challenging and agile locomotion \cite{ibarz2021train}. It doesn't require an accurate model of the system dynamics to achieve robust locomotion \cite{peng2020learning, miki2022learning, kumar2021rma}. To enable the emergence of agile locomotion gaits, it is necessary to carefully design a reward function \cite{kumar2021rma} that incentivizes the intended behavior and punishes undesired motions. This process \cite{hu2020learning} is laborious and time-consuming, and the resulting reward parameters can be brittle. Incorporating motion priors, such as modulating trajectory generators \cite{iscen2018policies}, can improve the overall robustness and learning stability \cite{lee2020learning, miki2022learning}. Another approach that produces natural-looking gaits is learning from reference motions \cite{peng2018deepmimic, ma2021learning}. Using information from motion capture data from real animals, the agent is conditioned to simultaneously imitate a reference clip and execute a goal-oriented task \cite{peng2018deepmimic, won2020scalable, escontrela2022adversarial, peng2021amp}.  

\par \emph{\textbf{Simulation-to-Real Policy Transfer}}. To ensure policies can be transferred to the real world (sim-to-real), domain randomization (DR) \cite{tobin2017domain} can be applied during training in simulation. By randomizing the physics parameters and the agent's properties during training, the policy becomes robust in a variety of environments \cite{peng2018sim}. Performance can be further improved by adapting the policies via system identification in the real-world setting\cite{peng2020learning, lee2020learning, tan2018sim}. A common approach is to encode the physics parameters into a latent representation \cite{peng2020learning, kumar2021rma, miki2022learning}. Then, trajectories sampled from the real-world can be used to search the latent space at deployment time \cite{peng2020learning}, or estimated from the recent history of states and actions \cite{kumar2021rma, miki2022learning}. In the context of policy transitions, previous work has shown great success in simulation \cite{soeseno2021transition}, but suffer from limitations in the real-world that we aim to address.

In our work, we develop locomotion policies that learn different gaits through motion imitation and can be deployed in a real-world setting via domain randomization. Distinctly from previous works, each gait of our library is learned as an independent locomotion policy. To expand the versatility of the overall locomotion, we employ a meta-controller that robustly unifies these independent policies through the use of our novel transition mechanism.

\begin{figure*}[t!]
\centering
\vspace{0.05cm}
\includegraphics[width=1.0\textwidth]{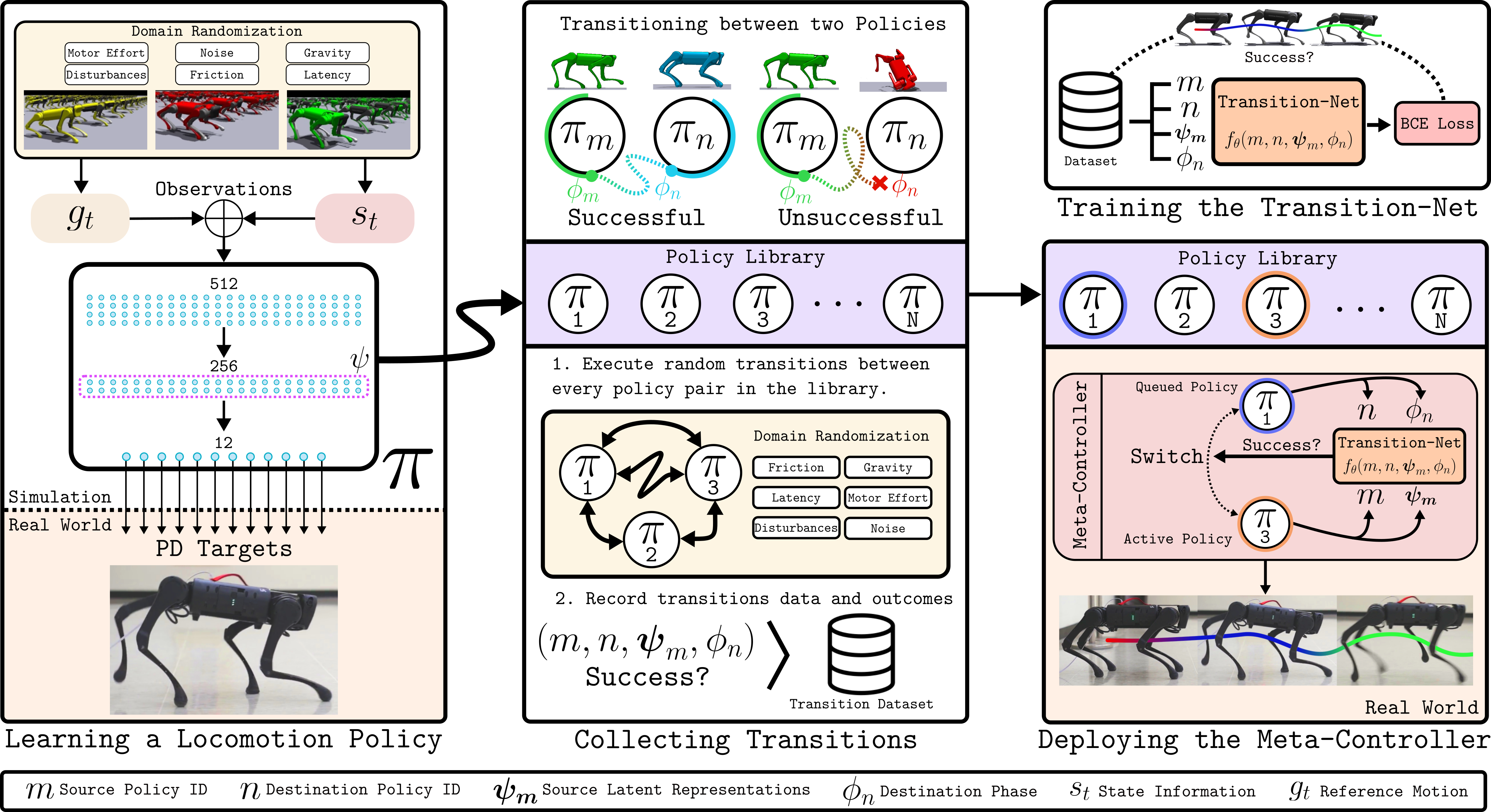}
\caption{We start by learning gait policies individually to build a policy library. Then, we collect transitions between every policy pair in the library. Our \textit{transition-net} acts as a robust transition strategy that enables the meta-controller to execute transitions in the real-world.}
\label{fig:diagram}
\end{figure*}

%% file: Sections/03_method.tex
\section{Methodology}
Our goal is to enable real-world quadruped robots to incrementally expand their library of locomotion gaits without altering previously learned ones. For this purpose, we contain the complexity of the gaits by training independent policies that specialize in a particular gait. Then, we construct a library of robust policies that can transfer to real-world robots with the use of domain randomization. Similar to \cite{soeseno2021transition}, we introduce a transition mechanism to link the independent policies by instantaneously switching between any two arbitrary policies. But, where previous works used phase to identify configurations that yield successful outcomes, instead, we propose a transition strategy that uses the latent representations of the policies, the \textit{transition-net}. During deployment with a real-world A1 quadruped robot, we construct a meta-controller that can access all policies available in the library and regulates the switch between an active and a queued policy.

\begin{table}[b!]
    \centering
    \caption{Parameters for domain randomization with uniform sampling.}
    \label{tab:dr_params}
    \begin{tabular}{ |l|c|c| } 
        \hline
        \textbf{Parameter} & \textbf{Range} & \textbf{Type} \\ \hline
        Gravity & [0.8, 1.2] & Scaling \\ \hline
        Action Noise & [-0.03, 0.03] & Additive \\ \hline
        Observation Noise & [-0.03, 0.03] & Additive \\ \hline
        Rigid Bodies Mass & [0.85, 1.15] & Scaling \\ \hline
        Ground Friction & [0.25, 1.5] & -- \\ \hline
        Observation Latency & [0.0, 0.020] s  & -- \\ \hline
        Stiffness Gain (PD Controller) & [45, 75] & -- \\ \hline
        Damping Gain (PD Controller) & [0.9, 1.8] & -- \\ \hline
    \end{tabular}
\end{table}

\subsection{Learning Independent Policies for Quadruped Robots}
\par Each policy in our skill repertoire learns a locomotion gait from a reference clip using a  motion imitation framework in simulation\cite{peng2018sim, peng2019mcp, peng2020learning}. Specifically, a policy $\pi$ is learned by maximizing the expected return,

\begin{equation}
    \label{eq:ppo_objective}
    J(\pi) =\mathbb{E}_{\tau \sim p(\tau|\pi)}[\sum_{t=0}^{T}\gamma^{t}r_t] \,,
\end{equation}

\noindent where $p(\tau|\pi)$ is the likelihood of a trajectory $\tau$ given the policy $\pi$, and $\sum_{t=0}^{T}\gamma^{t}r_t$ is the accumulated reward collected during the trajectory. $r_{t}$ denotes the reward collected at time $t \in T$, where $T$ denotes the length of each episode, and $\gamma \in [0, 1]$ represents the discount factor for future rewards. We train policies similar to \cite{lee2020learning,peng2018deepmimic}, where the policy learns an action distribution by imitating a reference motion clip. The input of the policy consists of the agent's state $\boldsymbol{s}_t$ and data from the reference motion clip $\boldsymbol{g}_t$. The policy is modeled as a feed-forward network that outputs the action distribution $\boldsymbol{a}_t$ given the current state and reference motion data $\pi(\boldsymbol{a}_t \vert \boldsymbol{s}_t, \boldsymbol{g}_t)$. See Section \ref{sec:implementation_details} for more details.

\par To transfer to the real-world, we apply extensive domain randomization (DR) of the simulation physics parameters and add other disturbances during the training process. This improves the inherent robustness of each policy and minimizes the performance gap between the simulation and the real-world setting\cite{li2021reinforcement, siekmann2021sim, lee2020learning}. With careful selection and application of the DR parameters described in Table~\ref{tab:dr_params}, we can reliably perform sim-to-real deployment of the independent locomotion policies with no failures (see Figure \ref{fig:teaser}). For each gait, the training process is realized independently and from scratch. At this point, we have a collection of independent policies that are robust enough to be deployed to real-world robots. However, each policy only enables the robot to perform one specific gait, and a method to transition between them has to be established.

\subsection{Transitioning between Policies with Transition-Net}
\par Each policy is considered a robust periodic controller capable of recovery from unstable states within an unspecified tolerance. For example, when the agent stumbles due to external perturbations, the policy acts to prevent its collapse and resumes its normal periodic motion afterward. Given this behavior, when a policy from the library is executing, it is possible to instantaneously switch the execution to a different policy at a particular destination phase. With proper timing of the switch and a  good choice of the destination phase, the new active policy takes control, corrects unstable behavior due to the initial configuration, and resumes its normal periodic execution.

Our method provides a way to consistently identify transition configurations that yield successful outcomes, i.e., where the agent remains stable after a transition. It takes advantage of the rich latent representations of each policy, obtained from the last hidden layer. We formulate a transition function $f_\theta(\cdot)$ that maps the transition configurations to their outcome, where $\theta$ represents the weights of a feed-forward neural network, called the \textit{transition-net}. The transition outcome is denoted by the binary variable $\alpha \in \{0, 1\}$, and the transition configuration $\mathcal{C}$ is represented as a tuple of four elements,

\begin{equation}
    \mathcal{C} = (m, n, \boldsymbol{\psi}_m, \phi_n),
\end{equation}

\noindent where $m$ and $n$ are identifiers of the source and destination policy, respectively; $\boldsymbol{\psi}_m$ is a high-dimensional vector denoting the latent representations of the source policy (see Figure \ref{fig:diagram}), and $\phi_n \in [\,0, 1)$ is the phase of the destination policy. 

To train the \textit{transition-net}, we collect millions of transition samples in simulation as detailed in Section \ref{sec:implementation_details}. Using this transition dataset, we train the \textit{transition-net} in a supervised manner solving for a binary classification problem, where it aims to predict whether a transition configuration $\mathcal{C}$ would result in a successful outcome $\alpha=1$ or a failure $\alpha=0$. It optimizes for the binary cross-entropy (BCE) loss with $\alpha$ as the classification label $y$,

\begin{equation}
    \label{eq:bce}
    y log(\hat{y}) + (1-y) log(1-\hat{y}),
\end{equation}

\noindent where $y$ denotes the recorded ground truth outcome, and $\hat{y}$ are the network's predictions. 

\subsection{Unifying Policies with a Meta-Controller}
\label{sec:meta_policy}

\begin{algorithm}[t!]
    \vspace{0.05cm}
    \caption{Meta-Controller with \textit{Transition-Net} Strategy}
    \label{alg:meta_controller_pipeline}
    \begin{algorithmic}
        \State \textbf{Input}: Policy library $\Pi$ and Transition-Net $f_\theta(\cdot)$
        \State $\pi_m \gets \text{sample a policy from policy library } \Pi$
        \ForEach {time step $t$ $\in$ Real World}
            \State $(\boldsymbol{a}_t\,, \boldsymbol{\psi}_m) \sim \pi_m(\boldsymbol{s}_t, \boldsymbol{g}_t)$ \Comment{Update step.}
            \State PDTargetControl($\boldsymbol{a}_t$)  \Comment{Controls the robot.}
            \State $\pi_n \gets$ RequestChangePolicy()
            \If{$\pi_m \neq \pi_n$}
                \State $\hat{y} \gets \max(\{f_\theta(m, n, \boldsymbol{\psi}_m, \phi_n); \phi_n \in [0, 1)\})$  
                \If {$\hat{y} > th$}
                    \State $\pi_m \gets \pi_n$   \Comment{Switches policy.}
                \EndIf
            \EndIf
        \EndFor
    \end{algorithmic}
\end{algorithm}

Next, to coherently unify all policies during deployment, we construct a meta-controller using the \textit{transition-net} estimates as transition scores. The meta-controller queries the \textit{transition-net} to identify the best transition configurations. It is responsible for choosing, scheduling, and executing the policies deployed on the robot using these elements:

\begin{itemize}
\item an active policy, which controls the robot by generating joint target angles actuated via PD control; 
\item a queued policy, to be switched for the active policy as soon as possible; and 
\item a transition function $f_\theta(\mathcal{C})$ that provides a score given the current configuration $\mathcal{C}$.
\end{itemize}

\par During runtime, we start by defining an initial active policy that controls the robot. The active policy can be initialized to any policy available in the library. At some point in time, a request for a policy change happens, and a different policy from the library is queued. Once a policy is queued, the meta-controller recognizes that a switch should happen. At every timestep (30Hz), it queries the transition function $f_\theta(\mathcal{C})$ and computes the transition score of switching the active policy for the queued policy. Note that we search over multiple destination phases $\phi_n$ when querying and choose the highest-scoring one. When the transition score crosses a predefined threshold, the queued policy becomes active and takes control of the robot. The meta-controller pipeline is presented in Algorithm~\ref{alg:meta_controller_pipeline}.

%% file: Sections/04_implementation_details.tex
\section{Implementation Details}
\label{sec:implementation_details}
\begin{figure*}[ht!]
    \vspace{0.05cm}
    \centering
    \includegraphics[width=1.0\textwidth]{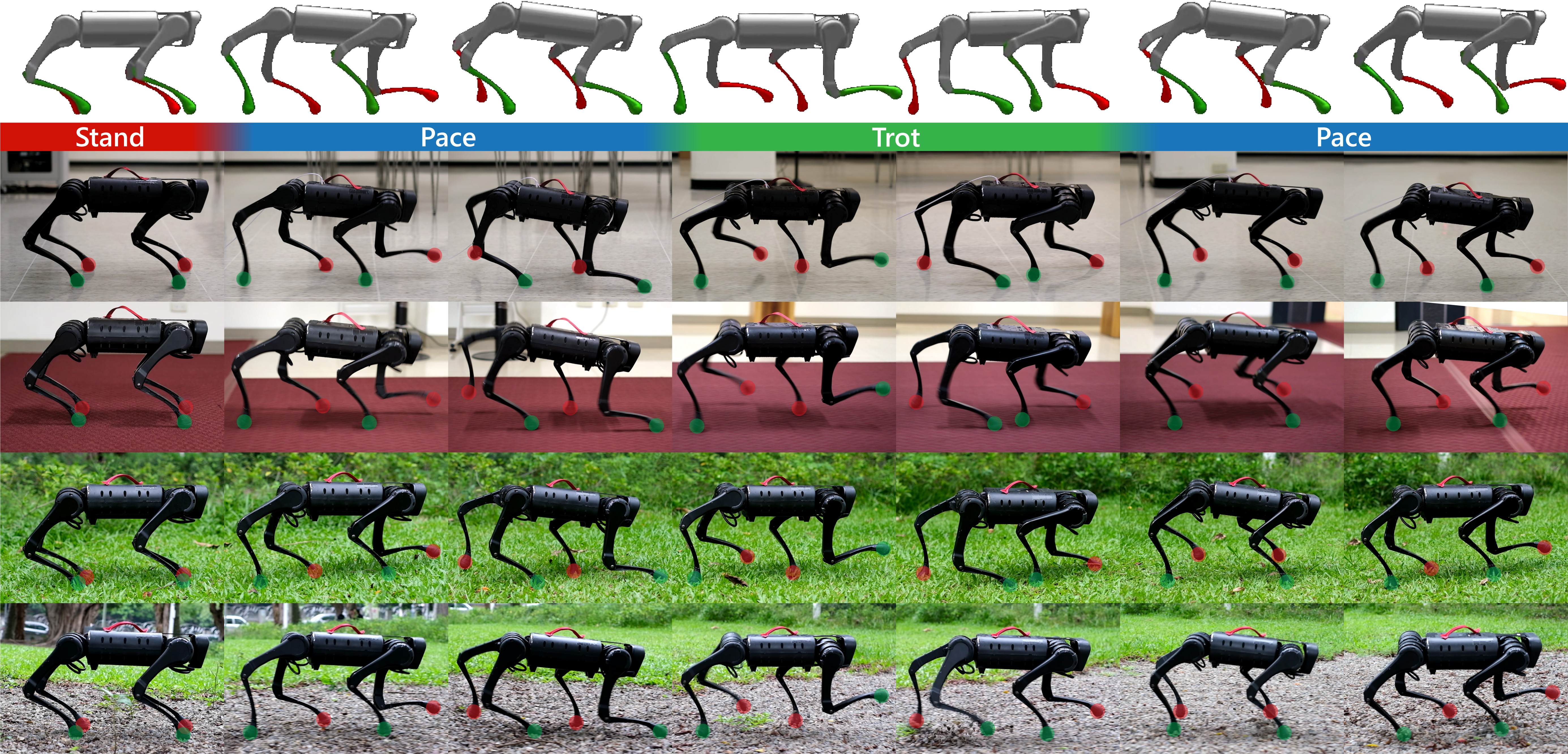}
    \caption{Transition between locomotion policies in simulation (first row) and in the real-world. Our policies are robust and can be deployed in a variety of environments. For more comprehensive results, please refer to the supplementary video.}
    \label{fig:transition_exp}
\end{figure*}
\subsection{Quadruped Robot and Simulation Engine}

In our experiments we use the Unitree A1 quadruped robot, which has 12 joints actuated via a PD controller. In the training stage, the simulated agent matches the configurations and properties of the real robot. The observation space of the policy is composed of the state $\boldsymbol{s}_t$ and the reference motion $\boldsymbol{g}_t$. The state $\boldsymbol{s}_t \in \mathbb{R}^{102}$ includes state information from the current and past two timesteps of the agent. A single state is composed of 12 joint angles, the orientation and angular velocities (6), a binary indicator of the contact for each foot (4), and the previous actions of the policy (12). The reference motion $\boldsymbol{g}_t \in \mathbb{R}^{56}$ contains the target poses from the motion capture data at 4 future timesteps, up to one second in the future \cite{lee2020learning}. It functions as an implicit phase variable by modulating the near future targets of the agent. When searching for the best destination phase of the queued policy, we shift the queued reference motion data in the time axis.

The gait policies are trained following the imitation learning objective from \cite{peng2018deepmimic}, with the PPO clip loss parameters described in Table~\ref{tab:training_hyperparams}. We developed the RL training environment using the simulator \textit{Isaac Gym} \cite{makoviychuk2021isaac}, which can accelerate the training by instancing several parallel environments (4096) in a single physics scene and exposing the simulation states via a \textit{PyTorch}-like API. With our implementation, the training process of a single locomotion policy takes less than one hour of wall clock time on a system equipped with an Intel i7-11800H and an RTX 3070 8GB. The policies can be deployed with the real robot in a zero-shot manner.

\begin{table}[t!]
    \centering
    \caption{Hyperparameters for training a locomotion policy with PPO.}
    \label{tab:training_hyperparams}
    \begin{tabular}{ |l|c| } 
        \hline
        \textbf{Parameter}& \textbf{Value} \\ \hline
        Number of Environments & 4096 \\ \hline
        Sequence Length & 24 \\ \hline
        Sequences per Environment & 4 \\ \hline
        Policy Optimization Iterations & 5 \\ \hline
        PPO Batch Size & 12288 \\ \hline
        Adam Optimizer LR & 3 x $10^{-4}$ \\ \hline
        Discount factor $\gamma$ & 0.95 \\ \hline
        Generalized Advantage Estimation $\lambda$ & 0.95 \\ \hline
        PPO Clip threshold & 0.2 \\ \hline
        KL threshold & 0.008 \\ \hline
        Entropy coefficient & 0.0 \\ \hline
    \end{tabular}
\end{table}

\subsection{Training of the Transition-Net}
Our proposed \textit{transition-net} is trained in a supervised manner with millions of transition samples collected from simulation with domain randomization as described in Table~\ref{tab:dr_params}. The samples contain paired labels of transition configuration and their corresponding outcome, where the source and destination policies are uniformly sampled from the library. Since the random switching strategy used to collect the samples introduce an imbalanced number of failure and successful cases, we sub-sample transition samples such that the number of success and failure samples are balanced. 

The \textit{transition-net} is implemented as a feed-forward network with $128-64-32$ neurons as the intermediate layers, with dropout ($p=0.4$) and ReLU activation functions applied after each layer except for the output layer, which uses a sigmoid. The neural network is trained for 100 epochs using a mini-batch of 128 samples, and the AdamW optimizer \cite{loshchilov2017decoupled} with a learning rate of $5\mathrm{e}{-4}$. During deployment, we use the output of the \textit{transition-net} as a scoring function, and only perform transitions when the score crosses the threshold $f_\theta(\mathcal{C}) > th$; where $th=0.95$.

\color{black}

%% file: Sections/05_experiments.tex
\section{Experiments}
To evaluate the effectiveness of our approach in identifying transition configurations that map to successful outcomes, we construct a policy library consisting of three locomotion gaits, namely, $pace$, $trot$, and $stand$. By using a static gait ($stand$) and two dynamic gaits ($pace$ and $trot$) we are able to run experiments that include dynamic-to-dynamic transitions as well as static-to-dynamic and vice-versa. Next, we unify these policies using a meta-controller that uses the \textit{transition-net} to score transition configurations and regulate the timing of a policy switch. Figure~\ref{fig:transition_exp} provides a visual illustration of the deployed meta-controller in the real-world. The difference between the $pace$ and $trot$ gaits is highlighted by the feet contact pattern, where $trot$ crosses the front and back legs while moving, enabling the robot to move at a faster speed. Notably, the policies can adapt to the different environments shown in the figure without requiring any additional tuning process in the real-world. In addition to the qualitative analysis of the policies, we also evaluate the transition robustness produced by our transition strategy compared to other existing works in Section \ref{sec:transition-robustness}. Finally, in Section \ref{sec:latent-representation} we investigate alternative input representations for the \textit{transition-net}, validating our design choices.

\subsection{Transition Robustness in Simulation and Real-World}
\label{sec:transition-robustness}
To evaluate the robustness of the transitions generated by the \textit{\textbf{transition-net (Ours)}}, we compare it to four other transition strategies: \textit{\textbf{random}}, \textit{\textbf{value function}} \cite{peng2018sim}, \textit{\textbf{transition motion tensor (TMT)}} \cite{soeseno2021transition}, and \textit{\textbf{TMT-value}}. \textit{\textbf{TMT-value}} is an extension of TMT we developed by integrating value function estimates into the querying function of TMT. We first evaluate the success rate of the transition strategies under increasing disturbance forces in simulation. Each strategy performs every possible transition pair while disturbance forces are applied at random directions and intervals (0.1 to 0.3 seconds). The disturbance magnitude starts at 0 N and increases up to 500 N in increments of 50 N. We collect 5000 transition samples for every combination of transition strategy, transition pair, and increment of disturbance magnitude.

\begin{figure}[t!]
    \centering
    \vspace{0.05cm}
    \includegraphics[width=0.49\textwidth]{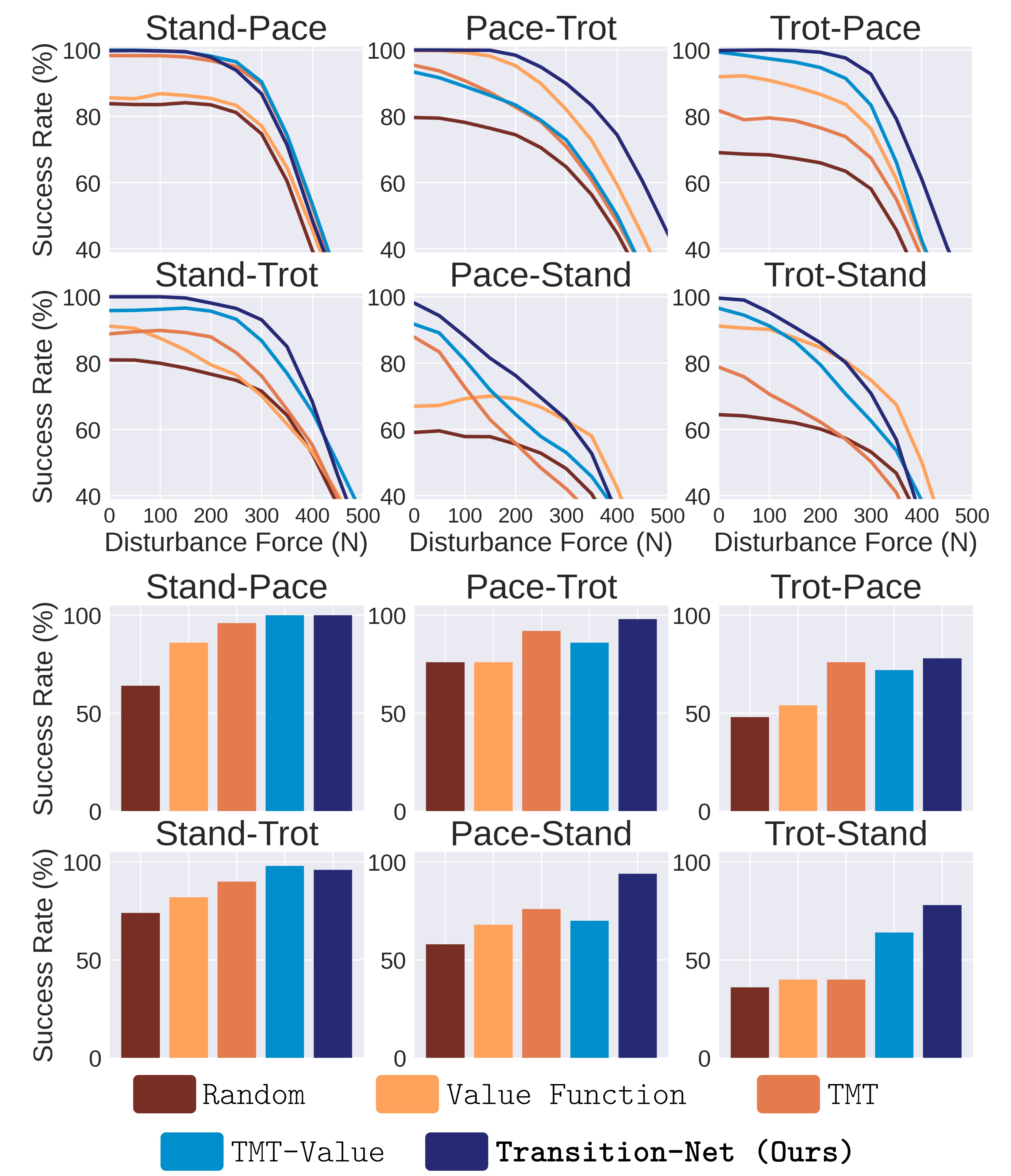}
    \caption{Impact of varying disturbances to the transition success rate in simulation (top) and real-world 
    transition success rate (bottom).}
    \label{fig:transition_success}
\end{figure}

Figure~\ref{fig:transition_success} (top) summarizes the success rate of each strategy and transition pair. As expected, randomly switching between policies without any heuristics yields suboptimal and unstable transitions. By employing the timing heuristics from the phase variable, \textit{\textbf{TMT}} achieves a success rate of 95\% for $pace$--$trot$. While the phase variable is a considerable indicator of when to perform a transition, it still fails to disambiguate between different states with similar phase values, e.g., from external disturbances or other motion inconsistencies. Other strategies such as \textit{\textbf{value function}}, \textit{\textbf{TMT-value}}, and \textit{\textbf{transition-net}} achieve a higher degree of robustness, highlighted by the higher success rate under the more intense disturbances. This is because more dynamic representations can better capture the state of the robot in relation to its environment dynamics. However, since the policy's value function is not specifically designed to facilitate transitions, \textit{\textbf{value function}} and \textit{\textbf{TMT-value}} strategies suffer from inconsistencies with lower success rates compared to our approach. 

\par Next, we deploy the meta-controller with the real A1 quadruped robot and once again evaluate the success rate of each transition strategy. We collect 50 transition samples for every possible combination of transition strategy and pair from the library. From Figure \ref{fig:transition_success} (bottom), we can observe that the robustness achieved in the simulation translates fairly well to the real-world setting, albeit with a lower overall success rate. The \textbf{\textit{value function}} strategy produces inconsistent transitions. This is expected as the value function is trained in the simulation setting and might produce noisy value estimates in the real-world setting. In this experiment we do not introduce external disturbances to the robot, allowing \textit{\textbf{TMT}} and \textit{\textbf{TMT-value}} to achieve high success rates and showing that phase serves as a reasonably accurate proxy. Finally, using the policy's latent representation with our \textit{\textbf{transition-net}} yields the most robust transitions. It can accurately identify the most appropriate configuration to perform the transitions for every pair in the real-world.

Jointly analyzing the results from the simulation and real-world settings, we observe that the most challenging transition are dynamic-to-static transitions, i.e., when the robot performs a full-stop after a moving gait, such as $pace$--$stand$ and $trot$--$stand$. This is evidenced by the steeper curves in Figure~\ref{fig:transition_success} (top) and the low success rate of the \textbf{\textit{random}} approach in the real-world in Figure~\ref{fig:transition_success} (bottom). The \textit{\textbf {transition-net}} achieves 19\% higher average success rate compared to the second best strategy, \textit{\textbf{TMT-value}}.

\subsection{Latent Representation of Agent State}
\label{sec:latent-representation}
To investigate the effectiveness of using the latent state representations $\boldsymbol{\psi}_m$ as the input of \textit{transition-net}, we experiment with two alternative representations: phase variable $\phi_m$ and the explicit state of the agent, a subset of $\boldsymbol{s}_t$. For each representation, we measure the accuracy of the \textit{transition-net} in predicting successful transition configurations given a balanced validation set. The dataset contains 5500 samples for each pair, with a total of 33000 samples. The experiment results shown in Table \ref{tab:accuracy_with_different_inputs} illustrate the strengths of each representation. 

The phase variable remains a solid low-dimensional proxy for the agent's state when the source policy is not very dynamic. This behavior is highlighted by the high accuracy in transitions where $stand$ is the source policy, but poor accuracy for all other transitions. A more descriptive state representation is to explicitly use the recent history of the agent's joint positions, orientation, angular velocity, and feet contacts. With this input, the \textit{transition-net} can internally estimate the phase information while having a more comprehensive state representation, resulting in higher accuracy than simply using the phase variable. In the end, the latent state representation encapsulates all of these desirable properties with the addition of implicit encoding of the environment obtained during the training process in simulation. Therefore it yields the highest prediction accuracy in transitions involving more dynamic states, such as the ones originating from $pace$ and $trot$.

\begin{table}[t!]
    \centering
    \caption{Effects of different state representations for the transition-net including (S)tand, (P)ace, and (T)rot gaits.}
    
    \label{tab:accuracy_with_different_inputs}
    \begin{tabular}{ |c|c|c|c|c|c|c| } 
        \hline
        \multirow{2}{*}{\textbf{Representation}}     & \multicolumn{6}{c|}{\textbf{Accuracy (\%)}}                                                       \\ \cline{2-7}
        									& \textbf{S-P}   & \textbf{S-T}   & \textbf{P-T}   & \textbf{P-S}   & \textbf{T-P}   & \textbf{T-S}  \\ \hline

        Phase ($\phi_m$)                & 88.94 & \textbf{87.59} & 63.86 & 67.42 & 66.14 & 65.01 \\ \hline
        Explicit ($\boldsymbol{s}_t$)                & \textbf{89.15} & 87.5  & 81.26 & 78.01 & 76.16 & 71.85 \\ \hline
        Latent ($\boldsymbol{\psi}_m$) & 87.14 & 86.97 & \textbf{82.45} & \textbf{78.68} & \textbf{78.3}  & \textbf{75.49} \\ \hline
    \end{tabular}
\end{table}

%% file: Sections/06_conclusion.tex

\section{Conclusion}
We proposed a robust transition strategy that enables a meta-controller to expand the versatility of locomotion in a quadruped robot. Containing the complexity of each gait in independent policies is pivotal in making the training and sim-to-real process more tractable. The experiments demonstrated that the latent state representation served as a reliable proxy for the state of the robot and its environment dynamics. It enabled the \textit{transition-net} to robustly identify successful transition configurations. Our setup of independent policies paired with the meta-controller enables iterative skill expansion while preserving learned ones. Finally, our massively parallel implementation allowed us to extensively iterate through policies and design choices.

In the future, we plan to perform experiments with a larger number of motion policies to further validate the iterative expansion property of our method. Additionally, we want to investigate more refined mechanisms of policy switching, such as generating transition trajectories. Doing so, would enable transitions between policies that are very distinct. We believe our work opens up interesting research directions toward versatile and robust legged robots.